\documentclass[a4paper, 10pt, conference]{ieeeconf}

\IEEEoverridecommandlockouts

\usepackage{graphicx}
\usepackage{amsmath}
\usepackage{amssymb}
\usepackage{cite}
\usepackage{url}
\usepackage{tikz}
\usepackage{booktabs}
\usepackage{multirow}
\usepackage{algorithm}
\usepackage{algorithmic}
\usetikzlibrary{arrows.meta, positioning, shapes.geometric, calc}

\usepackage{eso-pic}
\AddToShipoutPictureBG*{\AtPageUpperLeft{\put(\LenToUnit{54pt},\LenToUnit{-34pt}){\parbox[t]{504pt}{\centering\footnotesize
Accepted for publication at the 2027 IEEE/SICE International Symposium on System Integration (SII).\\[1.5pt]
\copyright~2027 IEEE. Personal use of this material is permitted. Permission from IEEE must be obtained for all other uses, in any current or future media, including reprinting/republishing this material for advertising or promotional purposes, creating new collective works, for resale or redistribution to servers or lists, or reuse of any copyrighted component of this work in other works.}}}}

\newcommand{\cmark}{\checkmark}
\newcommand{\xmark}{$\times$}
\newcommand{\sys}{TASC}

\title{\LARGE \bf
\sys{}: Time-to-Collision Anticipation and Selective Crowd Attention\\
for Non-Holonomic Service Robots
}

\author{Mohammadhossein Malmir$^{1,*}$, Mohammadali Rahmati$^{1,*}$, Mohammad Reza Kolani$^{2}$, and Alois Knoll$^{1}$%
\thanks{$^{*}$These authors contributed equally. $^{1}$Chair of Robotics, Artificial Intelligence and Real-time Systems, Department of Computer Engineering, School of Computation, Information and Technology, Technical University of Munich, Munich, Germany (e-mail: hossein.malmir@tum.de; mohammadali.rahmati@tum.de; k@tum.de). $^{2}$Chair of Computing in Civil and Building Engineering, Georg Nemetschek Institute, Technical University of Munich, Germany (e-mail: mohammadreza.kolani@tum.de). This work was conducted in collaboration with NEURA Robotics.}%
\thanks{A preliminary version of this work was presented at the RSS~2026 Workshop on Human-Centric Mobile Manipulation~\cite{rahmati2026reinforcement} (non-archival). This paper adds a hardware-in-the-loop test, the full reward and attention specification, a re-evaluation in which every episode runs to its end, component tests, a retrained TTC ablation, and a corrected comparison with published results.}%
}

\begin{document}

\maketitle
\thispagestyle{empty}
\pagestyle{empty}

\begin{abstract}

Service robots that share corridors and lobbies with people must keep moving without endangering them. Reactive planners can stop indefinitely in dense crowds (the freezing robot problem), whereas the attention-based learned navigators that dominate the standard crowd benchmark assume holonomic motion or emit discrete actions unsuited to a differential-drive base. This paper presents \sys{}, a navigation system for a non-holonomic service robot that integrates an attention-based crowd encoder, an anticipatory collision feature, and a continuous-control policy. We adapt the dual-selection attention module, originally developed for discrete-action value learning, to proximal policy optimization for continuous velocity control, and add a closing-speed time-to-collision feature, a conservative estimate of the time to contact that serves as both an observation and a reward term. The policy is trained on a high-fidelity 3D robot model in NVIDIA Isaac Lab with 10{,}000 parallel environments, social-force-model pedestrians, and randomized crowd sizes. In 30{,}000 evaluation episodes, each run to its end, the policy timed out in fewer than 0.5\% at every tested crowd density. This liveness comes at a cost: pedestrian collisions rise from 4.5\% with five pedestrians to 17.4\% with ten. In an ablation with four seeds per variant, policies retrained without the time-to-collision feature succeed 4 to 7 percentage points less often. The policy runs in Gazebo without retraining and on a commercial differential-drive service robot through a ROS~2 package with an inference latency below 10\,ms, verified in a hardware-in-the-loop test with virtual pedestrians.

\end{abstract}

\begin{keywords}
Attention mechanisms, collision avoidance, deep reinforcement learning, mobile service robots, robot operating system, social robot navigation.
\end{keywords}

\section{INTRODUCTION}

Service robots are moving out of structured industrial cells and into airports, hospitals, shopping centers, and offices. There, a robot shares the floor with people who neither coordinate with it nor stand still, and navigation becomes a prerequisite for deployment: the robot must reach its goal safely and predictably, without making the people around it uncomfortable.

Classical planners handle this setting poorly because they treat people as moving obstacles, not as agents who react to the robot. In dense crowds, the result is unnecessary detours or a complete stop, known as the \emph{freezing robot problem}~\cite{trautman2010unfreezing}. Deep reinforcement learning instead learns interaction behavior directly, without a map~\cite{le2024comprehensive}, and recent architectures encode the relations between agents with graph networks and attention~\cite{chen2019crowd, chen2020relational, cheng2024multi}.

Two practical gaps separate this line of work from a deployable platform. First, the attention- and graph-based policies of the standard crowd benchmark~\cite{chen2019crowd, chen2020relational} are trained for a holonomic disk, whereas a service robot usually has a differential-drive base. Second, the methods that model a unicycle either select from a discrete action set~\cite{everett2021collision, cheng2024multi}, which produces a jerky velocity profile, or output continuous commands~\cite{xie2023drlvo, flogel2024socially} but do not select among pedestrians. Non-holonomic kinematics, continuous commands, and hardware deployment each exist in prior work, but not together with selective attention (Section~\ref{sec:positioning}). This paper addresses that combination as a system-integration problem rather than as an isolated algorithm. The target platform is MiPA (My Intelligent Personal Assistant), a commercial differential-drive service robot from NEURA Robotics~\cite{neura2024mipa}. We refer to the integrated system as \sys{} (Time-to-collision Anticipation and Selective Crowd attention).

Our contributions are of three types:

\begin{itemize}
    \item \textbf{Adapted method.} We adapt the dual-selection attention module (DSAM)~\cite{cheng2024multi}, designed for discrete-action value-based learning, to proximal policy optimization (PPO)~\cite{schulman2017proximal} for continuous non-holonomic velocity control, and pair it with a closing-speed time-to-collision (TTC) feature that enters both the observation and the reward.
    \item \textbf{System integration.} We implement the complete pipeline from training to hardware: a high-fidelity 3D robot model in NVIDIA Isaac Lab~\cite{mittal2023orbit} with 10{,}000 GPU-parallel environments, social force model (SFM)~\cite{helbing1995social} pedestrians with stochastic stopping, crowd-size randomization, and a ROS~2 package that runs the policy on the physical robot in under 10\,ms per inference.
    \item \textbf{Empirical characterization.} We report the resulting behavior together with its main weakness: in 30{,}000 evaluation episodes, the policy times out in fewer than 0.5\% at every tested crowd density, but its pedestrian collision rate grows from 4.5\% to 17.4\% as the crowd grows from five to ten pedestrians. We quantify this trade-off between liveness (continued progress) and safety and test the TTC feature in a retrained ablation.
\end{itemize}

\section{RELATED WORK}

\subsection{Reactive Methods and the Freezing Robot Problem}

Reciprocal velocity obstacles~\cite{van2008reciprocal} and optimal reciprocal collision avoidance (ORCA)~\cite{vandenberg2011orca} are the standard reactive baselines. ORCA assumes \emph{reciprocity}, i.e., that each agent takes half of the responsibility for avoidance. Real pedestrians do not reliably do so, and ORCA then fails~\cite{chen2019crowd, liu2023graph}. The dynamic window approach (DWA)~\cite{fox1997dynamic} samples velocities within the kinematic limits but evaluates them against a static snapshot of the environment.

In dense crowds, a planner that does not model how people make room for the robot comes to judge every forward path unsafe, and the robot freezes or makes unnecessary maneuvers~\cite{trautman2010unfreezing}.

\subsection{Learning-Based Crowd Navigation}

CADRL~\cite{chen2017decentralized} learns a value network over robot--pedestrian pairs and thus scores each pedestrian in isolation. LSTM-RL~\cite{everett2021collision} handles a variable number of neighbors with a recurrent network, at the cost of an ordering bias. SARL~\cite{chen2019crowd} introduced self-attention over neighbors; because every detected person receives a nonzero weight, the policy also attends to people who are irrelevant to the current decision, which can cause hesitation~\cite{cheng2024multi}.

Graph-based formulations model the structure of the crowd explicitly. RGL~\cite{chen2020relational} applies graph convolutions to human--human and human--robot relations and plans several steps ahead, Liu et al.~\cite{liu2023graph} infer pedestrian intent within an attention-based interaction graph, and SafeCrowdNav~\cite{xu2023safecrowdnav} adds a collision-probability safety score. These methods reason over every node, which is informative but scales as $O(N^2)$, and every node keeps a nonzero weight. A graph-relational variant~\cite{liu2023grrl} reduces the cost with local relations but still weights every node that it retains.

MODSRL~\cite{cheng2024multi} addresses this selection problem with DSAM, which suppresses neighbors whose attention score falls below a statistical threshold (mean $\mu$, standard deviation $\sigma$): $\mu-\sigma$ in a selective layer and $\mu-3\sigma$ in an inclusive layer, whose outputs are fused. MODSRL models the robot as a unicycle but selects from 41 discrete actions (five speeds in each of eight orientations spread evenly over the full circle, plus a stop), which we replace with a continuous, bounded command.

Non-holonomic, continuous, and hardware-validated crowd navigation exists, mostly outside this attention-based line of work. SARL and RGL were demonstrated on wheeled robots despite holonomic training~\cite{chen2019crowd, chen2020relational}, and SA-CADRL on a differential-drive vehicle~\cite{chen2017socially}; LSTM-RL was trained with unicycle kinematics and run among pedestrians on a ground robot~\cite{everett2021collision}; and Liu et al.\ ran a holonomically trained policy on a TurtleBot among real people~\cite{liu2023graph}. The closest system-level precedent is DRL-VO~\cite{xie2023drlvo}, a PPO policy that outputs continuous, forward-only commands for a differential-drive base. It was trained with SFM pedestrians and a velocity-obstacle reward and run in real crowds without retraining, but it encodes pedestrians as kinematic grid maps rather than attending over them. SESN~\cite{tian2025preference} learns continuous velocity commands for a ground vehicle, conditioned on a user-defined preference vector, in a factory-like arena with static obstacles and pedestrians, and depends on automatic curriculum learning (52\% instead of 93\% success on its harder map without it). Fl\"ogel et al.~\cite{flogel2024socially} train a continuous-action PPO policy for a unicycle with a long short-term memory (LSTM) encoder and a reward on velocity deviations and personal-space violations, in simulation and without attention over neighbors. EmoBipedNav~\cite{zhu2026emobipednav} couples social navigation with full bipedal dynamics on a humanoid platform.

\subsection{Time-to-Collision as an Anticipatory Feature}

Karamouzas et al.~\cite{karamouzas2014universal} showed that the interaction energy between pedestrians follows a power law in the projected time to collision rather than in the separation distance, so TTC is close to the quantity to which people themselves respond. RGL~\cite{chen2020relational} shows that a two-step lookahead avoids collisions that its one-step policy runs into, and SafeCrowdNav~\cite{xu2023safecrowdnav} penalizes predicted collisions. We use the closing-speed form of TTC, which never overestimates the time to contact under constant velocities (Section~\ref{sec:ttc}).

\subsection{Positioning of This Work}
\label{sec:positioning}

Table~\ref{tab:method_comparison} compares the capabilities of these methods. It separates three properties that are easily conflated, namely non-holonomic kinematics in \emph{training}, \emph{continuous} commands, and deployment on \emph{hardware}, none of which is new on its own. We deliberately list no success rates there: the reported rates come from different kinematic models and pedestrian simulators, and Van Der Meer et al.~\cite{vandermeer2025comparative} find that the human motion model used in training changes the resulting policy markedly. Section~\ref{sec:results} lists the reported rates with their sources.

None of the surveyed methods combines selective attention, continuous non-holonomic control, and an explicit anticipatory feature, and the only one with selective attention (MODSRL) has not been run on hardware. \sys{} addresses this gap. Our contribution is the combination and its realization on a robot, not the invention of its components.

\begin{table}[tbp]
\caption{Capabilities of Crowd-Navigation Methods}
\label{tab:method_comparison}
{\centering\scriptsize
\begin{tabular}{|l|c|c|c|c|c|}
\hline
\textbf{Method} & \textbf{Non-holo.} & \textbf{Contin.} & \textbf{Sel.\ attn.} & \textbf{Anticip.} & \textbf{HW}\\
\hline
SA-CADRL~\cite{chen2017socially} & --- & \xmark & \xmark & \xmark & \cmark\\
\hline
LSTM-RL~\cite{everett2021collision} & \cmark & \xmark & \xmark & \xmark & \cmark\\
\hline
SARL~\cite{chen2019crowd} & \xmark & \xmark & \xmark & \xmark & \cmark\\
\hline
RGL~\cite{chen2020relational} & \xmark & \xmark & \xmark & \cmark & \cmark\\
\hline
SafeCrowdNav~\cite{xu2023safecrowdnav} & --- & \xmark & \xmark & \cmark & \xmark\\
\hline
Liu et al.~\cite{liu2023graph} & \xmark & \cmark & \xmark & \cmark & \cmark\\
\hline
DRL-VO~\cite{xie2023drlvo} & \cmark & \cmark & \xmark & \cmark & \cmark\\
\hline
MODSRL~\cite{cheng2024multi} & \cmark & \xmark & \cmark & \xmark & \xmark\\
\hline
SESN~\cite{tian2025preference}$^\dagger$ & \cmark & \cmark & \xmark & \xmark & \xmark\\
\hline
\textbf{\sys{} (ours)} & \cmark & \cmark & \cmark & \cmark & \cmark$^\ast$\\
\hline
\end{tabular}\par}
\vspace{1.2mm}
{\scriptsize Each row reflects the configuration reported in the cited work. Non-holo.: non-holonomic kinematics \emph{in training}. Contin.: continuous actions. Sel.\ attn.: attention that discards neighbors rather than down-weighting them. Anticip.: prediction, lookahead, or a forward-looking collision term in the observation or reward. HW: demonstrated on a physical robot. ---: not stated. $^\dagger$Factory-like benchmark rather than crowd navigation; depends on automatic curriculum learning. $^\ast$Hardware-in-the-loop: physical robot, virtual pedestrians, no perception (Section~\ref{sec:hil}).\par}
\end{table}

\section{METHOD}

\subsection{Problem Formulation}

We formulate navigation as a Markov decision process $\mathcal{M} = (\mathcal{S}, \mathcal{A}, \mathcal{P}, \mathcal{R}, \gamma)$ with state space $\mathcal{S}$, action space $\mathcal{A}$, transition kernel $\mathcal{P}$, reward function $\mathcal{R}$, and discount factor $\gamma$. We seek a policy that maximizes the expected discounted return. An episode ends when the robot reaches the goal, collides with a pedestrian or a wall, or exceeds the time limit $T$.

The task is a \emph{circle-crossing} scenario in a walled $10\,\text{m} \times 10\,\text{m}$ room: pedestrians start on a circle of radius 4.5\,m and walk to the diametrically opposite point and back, which makes crossing encounters the common case. The robot start, the goal, and the crowd size are randomized in every episode (Section~\ref{sec:randomization-config}).

\subsection{State Representation}

The robot state $\boldsymbol{s}_t^{\text{robot}} \in \mathbb{R}^7$ contains kinematic and goal-relative terms,
\begin{equation}
\boldsymbol{s}_t^{\text{robot}} = [d_g, \hat{g}_x, \hat{g}_y, v_{\text{pref}}, v_{\text{linear}}, \omega, r]^{\top} \text{,}
\end{equation}
where $d_g$ is the goal distance, $(\hat{g}_x, \hat{g}_y)$ the unit goal direction in the body frame, $v_{\text{pref}}=1.0$\,m/s the preferred speed, $v_{\text{linear}}$ and $\omega$ the current linear and angular velocities, and $r=0.35$\,m the robot radius.

Each pedestrian $i$ contributes an eight-dimensional vector
\begin{equation}
\boldsymbol{h}_t^{(i)} = [p_x^{(i)}, p_y^{(i)}, v_x^{(i)}, v_y^{(i)}, r_i, d_i, r_i + r, \text{TTC}_i]^{\top} \text{,}
\label{eq:ped_state}
\end{equation}
where $(p_x^{(i)}, p_y^{(i)})$ is its position relative to the robot and $(v_x^{(i)}, v_y^{(i)})$ its velocity, both expressed in the body frame, $r_i=0.3$\,m is the pedestrian radius, $d_i$ is the center-to-center distance, $r_i + r \equiv r_{\text{combined}}$ is the combined collision radius, and $\text{TTC}_i$ is the normalized time-to-collision defined in Section~\ref{sec:ttc}. Zero-mean Gaussian noise with a standard deviation of 0.1\,m is added to the observed positions, in training and in evaluation; $d_i$, the velocities, and $\text{TTC}_i$ are computed from the exact state. The observation $\boldsymbol{o}_t$ concatenates the robot state with the vectors of the $N$ observed pedestrians, padded to $N_{\max}=15$, so that $\dim(\boldsymbol{o}_t) = 7 + 15 \times 8 = 127$.

\subsection{Time-to-Collision Feature}
\label{sec:ttc}

Let $\boldsymbol{p}_{\text{rel}}^{(i)}$ and $\boldsymbol{v}_{\text{rel}}^{(i)}$ denote the position and velocity of pedestrian $i$ relative to the robot (its velocity minus that of the robot), and let $\hat{\boldsymbol{d}}_i = \boldsymbol{p}_{\text{rel}}^{(i)}/d_i$ be the unit vector from the robot to the pedestrian. The \textbf{closing-speed TTC} divides the remaining gap by the speed at which it shrinks,
\begin{equation}
\text{TTC}_{\text{raw}}^{(i)} = \frac{d_i - r_{\text{combined}}}{v_{\text{closing}}^{(i)}} \text{,}
\label{eq:ttc}
\end{equation}
where $v_{\text{closing}}^{(i)} = -\,\boldsymbol{v}_{\text{rel}}^{(i)} \cdot \hat{\boldsymbol{d}}_i$ is positive when the two approach each other.

\begin{figure}[tbp]
\centering
\begin{tikzpicture}[font=\footnotesize,
    rob/.style={circle, draw=blue!70, fill=blue!20, thick, minimum size=7mm, inner sep=0pt},
    ped/.style={circle, draw=red!70, fill=red!20, thick, minimum size=6mm, inner sep=0pt},
    vel/.style={-{Stealth[length=4pt]}, very thick}]

  \begin{scope}[shift={(0,0)}]
    \node[rob] (R) at (0.55,1.35) {\tiny R};
    \node[ped] (P) at (3.05,1.35) {\tiny P};
    \draw[vel, blue!70]  (R.east) -- ++(0.85,0) node[midway,above=0pt] {$\boldsymbol{v}_r$};
    \draw[vel, red!70]   (P.west) -- ++(-0.75,0) node[midway,below=0pt] {$\boldsymbol{v}_p$};
    \draw[gray, {Stealth[length=3pt]}-] (R.center) -- ++(145:0.35) node[above left=-3pt and -2pt,gray] {$r$};
    \draw[gray, {Stealth[length=3pt]}-] (P.center) -- ++(35:0.30) node[above right=-3pt and -2pt,gray] {$r_i$};
    \draw[<->, thick] (R.center|-0,0.55) -- (P.center|-0,0.55) node[midway,fill=white,inner sep=1pt] {$d_i$};
    \draw[gray!60, dashed] (R.center) -- (R.center|-0,0.5);
    \draw[gray!60, dashed] (P.center) -- (P.center|-0,0.5);
    \node[font=\scriptsize\bfseries] at (1.8,2.25) {(a) head-on};
  \end{scope}

  \begin{scope}[shift={(4.05,0)}]
    \node[rob] (R2) at (0.35,0.45) {\tiny R};
    \node[ped] (P2) at (2.80,0.45) {\tiny P};
    \coordinate (X) at (1.55,1.55);
    \draw[gray!55, dashed] (R2.center) -- (X);
    \draw[gray!55, dashed] (P2.center) -- (X);
    \draw[vel, blue!70] (R2.center) -- ($(R2.center)!0.55!(X)$);
    \node[blue!70, left=1pt] at ($(R2.center)!0.42!(X)$) {$\boldsymbol{v}_r$};
    \draw[vel, red!70]  (P2.center) -- ($(P2.center)!0.55!(X)$);
    \node[red!70, right=1pt] at ($(P2.center)!0.42!(X)$) {$\boldsymbol{v}_p$};
    \draw[orange!85, dotted, thick] (X) circle (0.42);
    \fill[orange!85] (X) circle (0.05);
    \draw[gray, {Stealth[length=3pt]}-] (X) -- ++(48:0.42);
    \node[gray, right=0pt, font=\scriptsize] at ($(X)+(48:0.42)$) {$r_{\text{combined}}$};
    \node[font=\scriptsize\bfseries] at (1.55,2.25) {(b) crossing};
  \end{scope}
\end{tikzpicture}
\caption{Closing-speed TTC. (a) Head-on approach: the estimate equals the time to contact. (b) Crossing paths: the separation reaches $r_{\text{combined}}$ later than estimated, if at all, so the feature is conservative.}
\label{fig:ttc}
\end{figure}
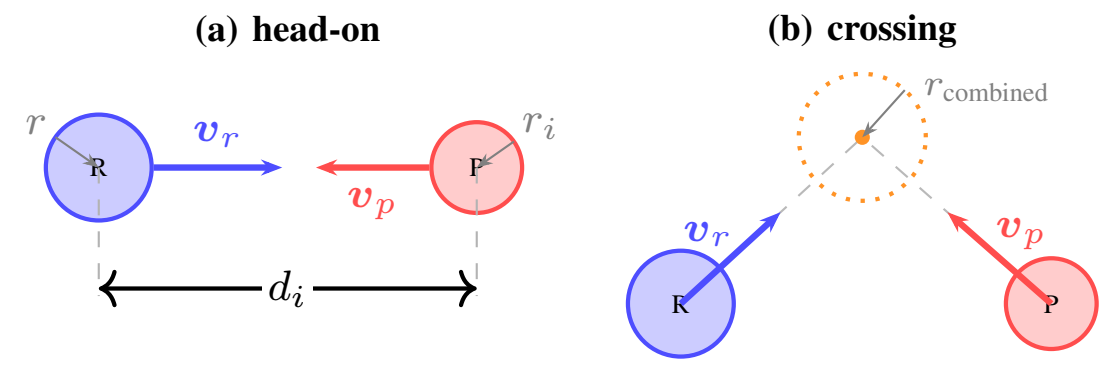

This value is a lower bound on the \textbf{geometric TTC}, the time $\tau$ at which the separation extrapolated at constant velocity, $\|\boldsymbol{p}_{\text{rel}}^{(i)} + \tau\, \boldsymbol{v}_{\text{rel}}^{(i)}\|$, reaches $r_{\text{combined}}$, because this separation is convex in $\tau$ and initially shrinks at the rate $v_{\text{closing}}^{(i)}$. The closing-speed TTC is therefore exact for a head-on approach and conservative otherwise (Fig.~\ref{fig:ttc}); it also responds to an approaching pedestrian whose extrapolated path would miss the robot. Our implementation takes the smaller of the two values, which by this bound is \eqref{eq:ttc} except in the degenerate case below.

The raw value is normalized by the horizon $T_{\text{TTC}}=5$\,s,
\begin{equation}
\text{TTC}_i = \min\!\bigl(\text{TTC}_{\text{raw}}^{(i)} / T_{\text{TTC}},\, 1\bigr) \in [0, 1] ,
\end{equation}
so that 0 denotes an imminent collision and 1 denotes no threat within the horizon.

\textbf{Degenerate cases.} If the pair is not approaching ($v_{\text{closing}}^{(i)} \leq 0.01$\,m/s), \eqref{eq:ttc} is not evaluated and $\text{TTC}_i = 1$; only within 5\,cm of contact can the geometric value then be smaller, and it is used. Inside the collision radius ($d_i < r_{\text{combined}}$), where the episode terminates, $\text{TTC}_i = 0$. No infinite or negative value therefore reaches the network.

\subsection{Action Space}

Actions are velocity commands for the differential-drive base,
\begin{equation}
\boldsymbol{a}_t = [v_{\text{linear}}, \omega]^{\top} \in [0,\, 1.0]\,\text{m/s} \times [-0.8,\, 0.8]\,\text{rad/s} .
\end{equation}
The policy outputs normalized actions in $[-1, 1]^2$, which are clipped and mapped linearly to these ranges.

The constraint $v_{\text{linear}} \geq 0$ excludes backward driving. This is a deployment requirement: the LiDAR and cameras of the platform face forward, so reversing would mean driving without sensor coverage. The cost is maneuverability in tight crowds; in-place rotation ($v_{\text{linear}} = 0$, $\omega \neq 0$) remains available.

\subsection{Reward Function}

Let $d_{\min} = \min_i d_i$ be the distance to the nearest pedestrian and $d_{\text{wall}}$ the distance from the robot center to the nearest wall. The step reward $R_t = \mathcal{R}(s_t, a_t)$ is a weighted sum of nine shaped terms, scaled by the control step $\Delta t = 1/60$\,s,
\begin{align}
R_t = \Delta t\,\bigl( & w_{\text{prog}} r_{\text{prog}} + w_{\text{coll}} r_{\text{coll}} + w_{\text{time}} r_{\text{time}} + w_{\text{smooth}} r_{\text{smooth}} \notag\\
& + w_{\text{bnd}} r_{\text{bnd}} + w_{\text{TTC}} r_{\text{TTC}} + w_{\text{goal}} r_{\text{goal}} \notag\\
& + w_{\text{fwd}} r_{\text{fwd}} + w_{\text{head}} r_{\text{head}} \bigr) ,
\end{align}
with weights $6.0$, $7.0$, $1.0$, $0.05$, $3.0$, $3.0$, $1.0$, $0.3$, and $1.0$ in this order, plus a terminal reward that is not scaled.

\textbf{Progress} gives a dense goal signal, $r_{\text{prog}} = k_{\text{prog}}\,(d_g(t-1) - d_g(t))$ with $k_{\text{prog}} = 5.0$, so that one meter of progress is worth $30\,\Delta t = 0.5$.

\textbf{Proximity and collision} are combined in one term,
\begin{equation}
r_{\text{coll}} = \begin{cases}
0, & d_{\min} > 2.0 \\
-(1 - \tilde{d})^2, & 0.65 \leq d_{\min} \leq 2.0 \\
-2.0, & d_{\min} < 0.65
\end{cases} \text{,}
\end{equation}
where $\tilde{d} = (d_{\min} - 0.65)/(2.0 - 0.65)$ is the normalized clearance, with $d_{\min}$ in meters and $0.65\,\text{m} = r_{\text{combined}}$. Because the episode terminates on contact, the third case applies only at the contact step.

\textbf{TTC anticipation} is active only under threat: with $\text{TTC}_{\min} = \min_i \text{TTC}_i$, $r_{\text{TTC}} = -(0.5 - \text{TTC}_{\min})$ if $\text{TTC}_{\min} < 0.5$ (i.e., below 2.5\,s) and 0 otherwise.

\textbf{Boundary} keeps the robot away from the walls: the term is zero beyond $d_{\text{soft}}=1.5$\,m, decreases linearly with scale $k_b = 0.5$ down to $d_{\text{hard}}=0.7$\,m, and is $-2.0$ inside $d_{\text{hard}}$. The episode is terminated as a wall collision at $d_{\text{wall}} = 0.55$\,m, a margin of 0.2\,m beyond the robot radius.

\textbf{Smoothness} penalizes sustained fast rotation rather than jerk: $r_{\text{smooth}} = k_\omega\,(\omega_{\text{th}} - |\omega|)$ if $|\omega| > \omega_{\text{th}}$ and 0 otherwise, with $k_\omega = 0.01$ and $\omega_{\text{th}} = 0.48$\,rad/s (60\% of the angular limit), so ordinary turning is not penalized.

\textbf{Forward motion} discourages stalling, $r_{\text{fwd}} = \min(v_{\text{linear}}, 0.5)$, and \textbf{heading} rewards moving while facing the goal, $r_{\text{head}} = \hat{g}_x \cdot \min(v_{\text{linear}}/v_{\text{pref}}, 1)$. \textbf{Goal} and \textbf{time efficiency} apply on success: $r_{\text{goal}} = 10.0$ once $d_g < 0.5$\,m, and $r_{\text{time}} = 1 - k_t\, t/T$ with $k_t = 0.1$; after the scaling by $\Delta t$, both are small next to the terminal reward.

\textbf{Terminal rewards} are $+50$ for reaching the goal and $-50$ for a pedestrian collision, a wall collision, or a timeout. A timeout thus carries the same penalty as a collision, although PPO bootstraps the value at a timeout; Section~\ref{sec:limitations} discusses the safety cost of this choice.

\subsection{Dual-Selection Attention Module}

DSAM~\cite{cheng2024multi} maps a variable number of pedestrians to a fixed-size crowd vector $\boldsymbol{K} \in \mathbb{R}^{50}$ by discarding neighbors rather than down-weighting them.

An embedding network $M_e$ computes joint robot--pedestrian embeddings $\boldsymbol{e}_i = M_e([\boldsymbol{s}_t^{\text{robot}}; \boldsymbol{h}_t^{(i)}]) \in \mathbb{R}^{50}$, and an interaction network $M_h$ maps them to $\boldsymbol{h}'_i = M_h(\boldsymbol{e}_i) \in \mathbb{R}^{50}$. Their means over the $N$ observed pedestrians, $\bar{\boldsymbol{e}} = \frac{1}{N}\sum_{i=1}^{N} \boldsymbol{e}_i$ and $\bar{\boldsymbol{h}}' = \frac{1}{N}\sum_{i=1}^{N} \boldsymbol{h}'_i$, summarize the crowd and serve as context for the scoring networks.

The \textbf{selective layer} computes scores $\alpha_i = M_\alpha([\boldsymbol{e}_i; \bar{\boldsymbol{e}}])$ and keeps only the set $\mathcal{S}_\alpha$ of pedestrians whose score is at least $\mu_\alpha - \sigma_\alpha$, where $\mu_\alpha$ and $\sigma_\alpha$ are the mean and standard deviation of $\{\alpha_i\}_{i=1}^{N}$ (with $\sigma_\alpha$ floored at 0.1); for normally distributed scores, this retains about 84\%. The softmax is taken over the kept set only, so a discarded pedestrian receives exactly zero weight (MODSRL assigns a negative constant in this layer instead):
\begin{equation}
\boldsymbol{C} = \sum_{i \in \mathcal{S}_\alpha} \alpha'_i\, \boldsymbol{h}'_i , \qquad \alpha'_i = \frac{\exp(\alpha_i)}{\sum_{j \in \mathcal{S}_\alpha} \exp(\alpha_j)} .
\end{equation}
The \textbf{inclusive layer} computes $\beta_i = M_\beta([\boldsymbol{h}'_i; \bar{\boldsymbol{h}}'])$ and keeps the set $\mathcal{S}_\beta$ with $\beta_i \geq \mu_\beta - 3\sigma_\beta$ (about 99.9\%), with weights $\beta'_i$ defined analogously:
\begin{equation}
\boldsymbol{D} = \sum_{i \in \mathcal{S}_\beta} \beta'_i\, \boldsymbol{e}_i .
\end{equation}
Because none of $N$ scores can lie more than $\sqrt{N-1}$ standard deviations below their mean, the inclusive layer discards no one when $N \leq 10$, as in our experiments, and then acts as ordinary soft attention.

\begin{figure}[tbp]
    \centering
    \begin{tikzpicture}[font=\footnotesize, node distance=0pt,
    bx/.style={rectangle, draw, rounded corners=1.5pt, align=center,
               minimum height=4.6mm, inner xsep=3pt, inner ysep=1pt},
    ar/.style={-{Stealth[length=4pt]}, thick}]

  \def\W{8.2}
  \def\P{3.9}

  \node[bx, fill=gray!14, minimum width=\W cm] (in) at (0,0)
      {Robot $\boldsymbol{s}^{\text{robot}}_t$ (7D) $+$ $N$ pedestrians ($8$D each)};
  \node[bx, fill=blue!14, minimum width=4.4cm, below=3.4mm of in] (me)
      {$M_e:\,[\boldsymbol{s};\boldsymbol{h}^{(i)}] \rightarrow \boldsymbol{e}_i$ \ (50D)};
  \node[bx, fill=blue!22, minimum width=4.4cm, below=3.0mm of me] (mh)
      {$M_h:\,\boldsymbol{e}_i \rightarrow \boldsymbol{h}'_i$ \ (50D)};
  \draw[ar] (in) -- (me);   \draw[ar] (me) -- (mh);

  \node[draw=red!55, fill=red!6, rounded corners=2pt, minimum width=\P cm,
        minimum height=2.00cm, anchor=north east] (L1) at ($(mh.south)+(-0.2,-0.62)$) {};
  \node[draw=blue!55, fill=blue!6, rounded corners=2pt, minimum width=\P cm,
        minimum height=2.00cm, anchor=north west] (L2) at ($(mh.south)+(0.2,-0.62)$) {};
  \node[font=\scriptsize\bfseries, text=red!65, anchor=north] at ([yshift=-1.2mm]L1.north) {selective: keep $\sim$84\%};
  \node[font=\scriptsize\bfseries, text=blue!65, anchor=north] at ([yshift=-1.2mm]L2.north) {inclusive: keep $\sim$99.9\%};

  \node[bx, fill=orange!18, minimum width=3.35cm] (a1) at ([yshift=-1.0mm]$(L1.north)!0.30!(L1.south)$)
      {score $\alpha_i = M_\alpha([\boldsymbol{e}_i;\bar{\boldsymbol{e}}])$};
  \node[bx, fill=orange!34, minimum width=3.35cm, below=5.0mm of a1] (c1)
      {$\boldsymbol{C}=\sum_i \alpha'_i\,\boldsymbol{h}'_i$};
  \node[bx, fill=orange!18, minimum width=3.35cm] (a2) at ([yshift=-1.0mm]$(L2.north)!0.30!(L2.south)$)
      {score $\beta_i = M_\beta([\boldsymbol{h}'_i;\bar{\boldsymbol{h}}'])$};
  \node[bx, fill=orange!34, minimum width=3.35cm, below=5.0mm of a2] (d2)
      {$\boldsymbol{D}=\sum_i \beta'_i\,\boldsymbol{e}_i$};

  \node[font=\scriptsize, anchor=east] at ([xshift=-1pt]c1.west) {};
  \draw[ar] (a1) -- node[right=0pt,font=\scriptsize,inner sep=1pt]{$\alpha_i\geq\mu_\alpha-\sigma_\alpha$} (c1);
  \draw[ar] (a2) -- node[right=0pt,font=\scriptsize,inner sep=1pt]{$\beta_i\geq\mu_\beta-3\sigma_\beta$} (d2);

  \draw[ar, red!55]  (mh.south) -- ++(0,-0.24) -| (L1.north);
  \draw[ar, blue!55] (mh.south) -- ++(0,-0.24) -| (L2.north);

  \node[bx, fill=green!25, minimum width=4.0cm, below=4.0mm of $(L1.south)!0.5!(L2.south)$] (fus)
      {$\boldsymbol{K}=(\boldsymbol{C}+\boldsymbol{D})/2 \in \mathbb{R}^{50}$};
  \draw[ar] (c1.south) |- ([yshift=2.0mm]fus.north) -| (fus.north);
  \draw[ar] (d2.south) |- ([yshift=2.0mm]fus.north) -| (fus.north);
\end{tikzpicture}
    \caption{DSAM. The selective layer scores on $\boldsymbol{e}_i$ and aggregates $\boldsymbol{h}'_i$; the inclusive layer does the reverse. $\alpha'_i$ and $\beta'_i$ are the masked softmax weights, and fusion yields the crowd vector $\boldsymbol{K}$.}
    \label{fig:dsam_architecture}
\end{figure}
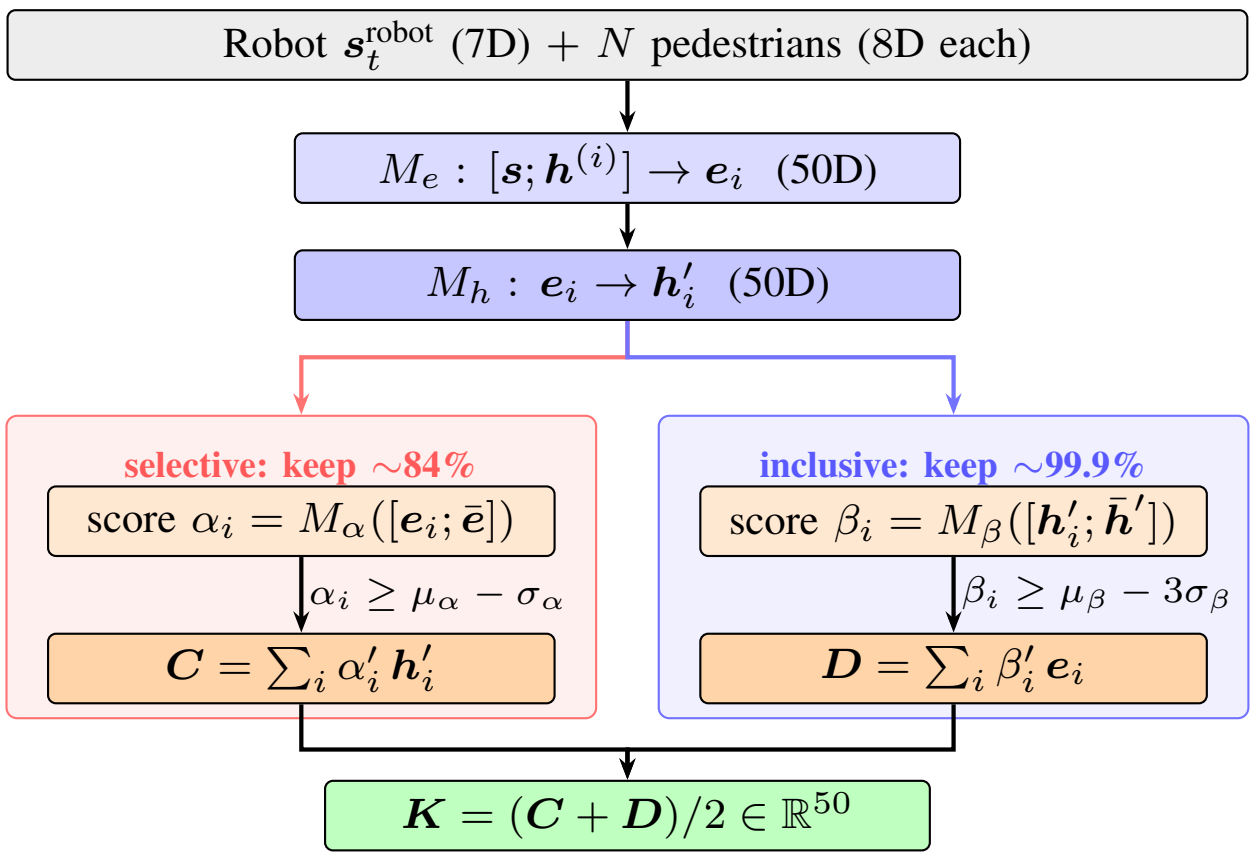

The layers aggregate crosswise: the selective layer scores on $\boldsymbol{e}_i$ but sums $\boldsymbol{h}'_i$, and the inclusive layer scores on $\boldsymbol{h}'_i$ but sums $\boldsymbol{e}_i$, so that the near-field threat estimate and the wider crowd context are read from different representations. The crowd vector is $\boldsymbol{K} = (\boldsymbol{C} + \boldsymbol{D})/2$, and the policy input is $\boldsymbol{x} = [\boldsymbol{s}_t^{\text{robot}}; \boldsymbol{K}] \in \mathbb{R}^{57}$. Fig.~\ref{fig:dsam_architecture} shows the module, and Algorithm~\ref{alg:dsam} gives the forward pass.

\begin{algorithm}[tbp]
\caption{DSAM Forward Pass}
\label{alg:dsam}
\begin{algorithmic}[1]
\small
\REQUIRE $\boldsymbol{s}^{\text{robot}}_t$; \ $\boldsymbol{h}^{(i)}_t$ for the $N$ observed pedestrians
\ENSURE Policy input $\boldsymbol{x} \in \mathbb{R}^{57}$
\STATE \textbf{if} $N = 0$ \textbf{then return} $\boldsymbol{x} \leftarrow [\boldsymbol{s}^{\text{robot}}_t; \boldsymbol{0}]$
\STATE $\boldsymbol{e}_i \leftarrow M_e([\boldsymbol{s}^{\text{robot}}_t; \boldsymbol{h}^{(i)}_t])$, \ $\boldsymbol{h}'_i \leftarrow M_h(\boldsymbol{e}_i)$ \ for $i = 1, \dots, N$
\STATE $\bar{\boldsymbol{e}} \leftarrow \frac{1}{N}\sum_i \boldsymbol{e}_i$, \ $\bar{\boldsymbol{h}}' \leftarrow \frac{1}{N}\sum_i \boldsymbol{h}'_i$
\STATE $\alpha_i \leftarrow M_\alpha([\boldsymbol{e}_i; \bar{\boldsymbol{e}}])$, \ $\beta_i \leftarrow M_\beta([\boldsymbol{h}'_i; \bar{\boldsymbol{h}}'])$ \ for all $i$
\STATE $\mathcal{S}_\alpha \leftarrow \{i : \alpha_i \geq \mu_\alpha - \max(\sigma_\alpha, 0.1)\}$
\STATE $\mathcal{S}_\beta \leftarrow \{i : \beta_i \geq \mu_\beta - 3\max(\sigma_\beta, 0.1)\}$
\STATE $\boldsymbol{C} \leftarrow \sum_{i \in \mathcal{S}_\alpha} \alpha'_i\, \boldsymbol{h}'_i$, \ $\boldsymbol{D} \leftarrow \sum_{i \in \mathcal{S}_\beta} \beta'_i\, \boldsymbol{e}_i$
\RETURN $\boldsymbol{x} \leftarrow [\boldsymbol{s}^{\text{robot}}_t; (\boldsymbol{C} + \boldsymbol{D})/2]$
\end{algorithmic}
\end{algorithm}

\subsection{Policy Optimization}
\label{sec:ppo}

We train with PPO~\cite{schulman2017proximal}, using its clipped surrogate objective with a clipping range of 0.2 and generalized advantage estimation (GAE) with $\lambda = 0.95$; the actor and the critic share the DSAM encoder. The actor, with parameters $\theta$, outputs the mean $\boldsymbol{\mu}_\pi = \pi_\theta(\boldsymbol{x})$ of a Gaussian over the normalized actions. Its standard deviation $\boldsymbol{\sigma}_\pi$ is a learned, state-independent parameter, initialized at 0.3 and constrained to $[0.1, 0.8]$. The critic, with parameters $\phi$, estimates the state value $V_\phi(\boldsymbol{x})$.

Three settings differ from common PPO defaults. The learning rate starts at $5 \times 10^{-5}$ and is adapted to a target KL divergence of 0.005, which keeps the policy updates small, as the selection thresholds depend on score statistics that shift with every update. The entropy coefficient is reduced to $0.001$ to limit the action noise, which is clipped at the velocity limits. The discount factor is raised to $\gamma=0.995$, an effective horizon of about 3.3\,s at the 60\,Hz control rate, of the same order as the 5\,s TTC horizon.

\section{SYSTEM IMPLEMENTATION}

\subsection{Simulation Architecture}

\begin{figure}[tbp]
\centering
\includegraphics[width=0.66\columnwidth]{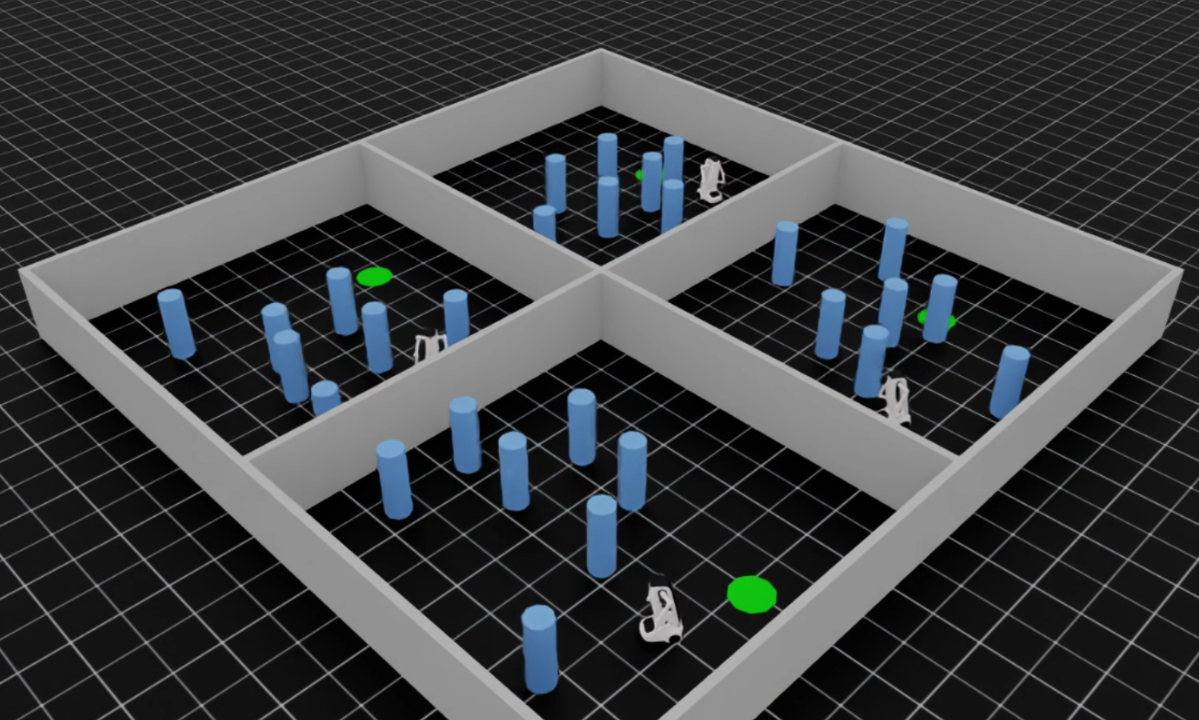}
\caption{Four of the 10{,}000 parallel training environments in Isaac Lab. Each contains the full 3D robot model, a crowd of randomized size (cylinders, collision response disabled), the goal marker, and the $10\,\text{m}\times10\,\text{m}$ room.}
\label{fig:envs}
\end{figure}

The environment is built on NVIDIA Isaac Lab~\cite{mittal2023orbit}, whose manager-based design keeps observations, actions, rewards, and terminations in independent modules: a task module handles goals, TTC, and observation assembly; a crowd manager runs the SFM; and a differential-drive action module drives the robot. Fig.~\ref{fig:envs} shows four environments. Three implementation decisions that do not follow from the algorithm were needed to keep 10{,}000 of them stable:

\begin{itemize}
    \item \textbf{Non-colliding pedestrians.} Physical collision response between the robot and the pedestrians is disabled, because contact-rich simulation at this scale destabilizes the GPU physics pipeline. Avoidance is therefore learned from the proximity and TTC rewards and the distance-based termination alone, and an episode ends as soon as $d_{\min} < r_{\text{combined}}$, so interpenetration cannot be exploited.
    \item \textbf{Base stability.} The center of mass of the robot model is 0.69\,m above its base link, which makes it prone to tipping. An angular damping of 5.0 and a linear damping of 1.0 prevent this without restricting the commanded velocity range.
    \item \textbf{Solver budgets.} The GPU buffers were sized explicitly ($2^{21}$ contacts, $2^{18}$ patches, 64\,MB heap) to prevent allocation failures during parallel rollouts.
\end{itemize}

\subsection{Training Pipeline}

\begin{figure}[tbp]
\centering
\begin{tikzpicture}[font=\footnotesize,
    bx/.style={rectangle, draw, thick, rounded corners=2pt, align=center,
               minimum height=6.5mm, minimum width=2.35cm, inner xsep=2pt, inner ysep=1pt},
    ar/.style={-{Stealth[length=4pt]}, thick}]

  \def\cA{1.25} \def\cB{4.10} \def\cC{6.95}
  \def\rA{0} \def\rB{-1.25} \def\rC{-2.45} \def\rD{-3.70} \def\rE{-4.90}

  \node[bx, fill=blue!14]   (scene) at (\cA,\rA) {Scene\\\scriptsize 10,000 envs};
  \node[bx, fill=orange!16] (crowd) at (\cB,\rA) {SFM Crowd\\\scriptsize Manager};
  \node[bx, fill=green!16]  (goal)  at (\cC,\rA) {Task\\\scriptsize goals, TTC};
  \draw[ar] (scene) -- (crowd);  \draw[ar] (crowd) -- (goal);

  \node[bx, fill=gray!12] (rs)  at (\cA,\rB) {Robot state (7D)\\\scriptsize$(d_g,\hat{\boldsymbol{g}},v,\omega,r)$};
  \node[bx, fill=gray!12] (ps)  at (\cB,\rB) {Ped.\ states (7D)\\\scriptsize$(\boldsymbol{p},\boldsymbol{v},r_i,d_i,r_i{+}r)$};
  \node[bx, fill=green!22] (ttc) at (\cC,\rB) {TTC\\\scriptsize per pedestrian};
  \draw[ar] (scene) -- (rs);  \draw[ar] (crowd) -- (ps);  \draw[ar] (goal) -- (ttc);
  \draw[ar, dashed, gray!70] (ps) -- (ttc);

  \node[bx, fill=gray!20, minimum width=7.95cm] (obs) at (\cB,\rC)
      {Observation $\boldsymbol{o}_t$ (127D)\ \ $=$\ \ robot (7D) $+$ $15 \times$ [ped.\ state $+$ TTC] (8D)};
  \draw[ar] (rs)  -- ([xshift=-2.6cm]obs.north);
  \draw[ar] (ps)  -- (obs.north);
  \draw[ar] (ttc) -- ([xshift=2.6cm]obs.north);

  \node[bx, fill=red!22]  (actor)  at (\cA,\rD) {Actor\\\scriptsize$(\boldsymbol{\mu}_\pi,\boldsymbol{\sigma}_\pi)$};
  \node[bx, fill=red!16]  (dsam)   at (\cB,\rD) {DSAM\\\scriptsize$\boldsymbol{K}$ (50D)};
  \node[bx, fill=cyan!16] (critic) at (\cC,\rD) {Critic\\\scriptsize$V_\phi$};
  \draw[ar] (obs.south) -- (dsam.north);
  \draw[ar] (dsam) -- node[above=1.5pt,font=\tiny,inner sep=0pt]{57D} (actor);
  \draw[ar] (dsam) -- node[above=1.5pt,font=\tiny,inner sep=0pt]{57D} (critic);

  \node[bx, fill=yellow!22] (act) at (\cA,\rE) {Diff.\ drive $(v,\omega)$};
  \node[bx, fill=cyan!22]   (ppo) at (\cC,\rE) {PPO update\\\scriptsize GAE $+$ clip};
  \draw[ar] (actor) -- (act);
  \draw[ar] (critic) -- (ppo);

  \draw[ar, dashed] (act.west) -- (-0.22,\rE) -- (-0.22,\rA) -- (scene.west);
\end{tikzpicture}
\caption{Training pipeline. The TTC is appended to each pedestrian state to form the 127D observation, which DSAM compresses to a 57D input for the actor and the critic; the actor outputs $(v, \omega)$.}
\label{fig:pipeline}
\end{figure}
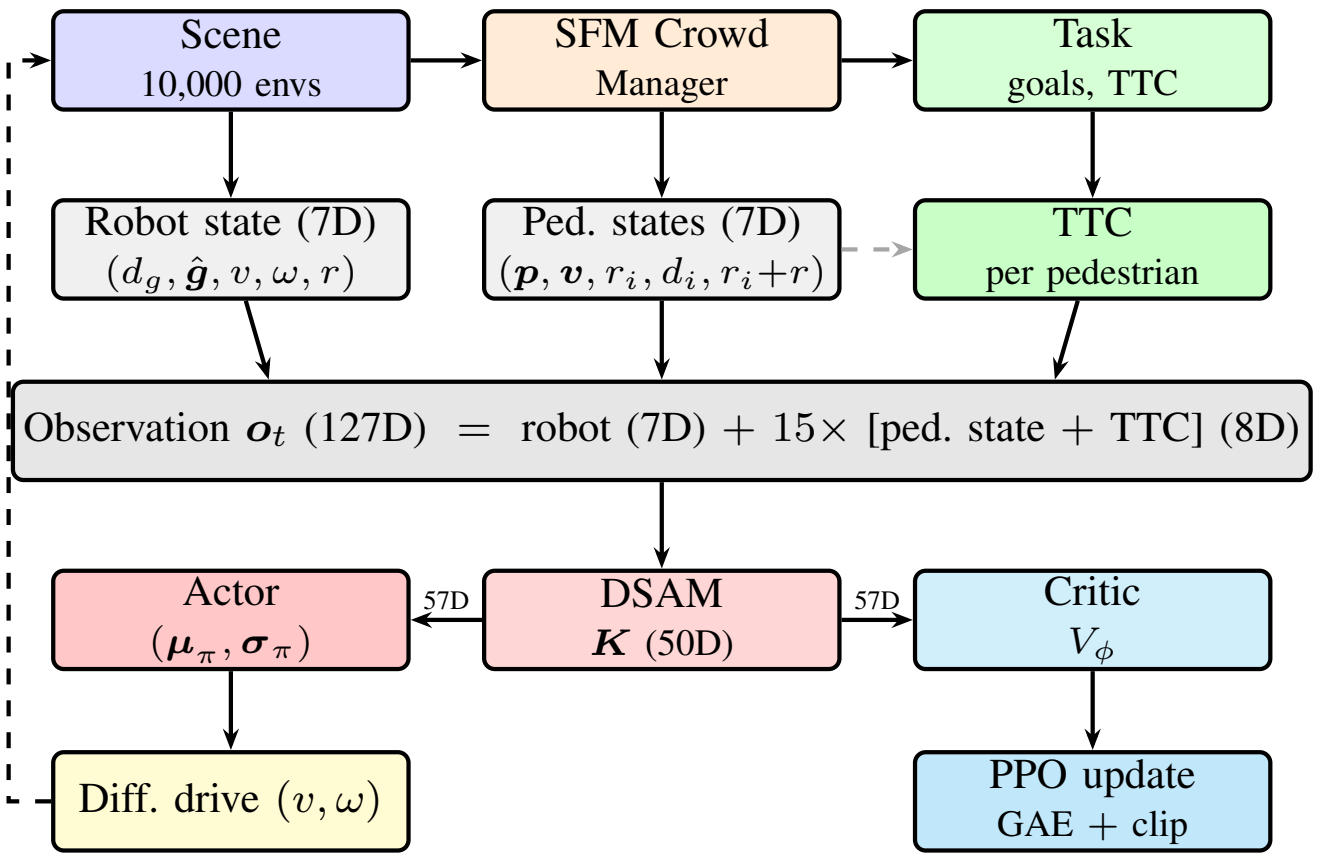

Fig.~\ref{fig:pipeline} shows the data flow from the parallel simulator to the velocity command. Each training iteration collects 32 steps in each of the 10{,}000 environments; eight PPO epochs then update the shared DSAM encoder, the actor, and the critic.

\subsection{Pedestrian Model}

Pedestrians follow a simplified SFM~\cite{helbing1995social}. The resultant of the social forces,
\begin{equation}
\boldsymbol{F}_i = \boldsymbol{F}_{\text{goal}}^{(i)} + \sum_{j \neq i} \boldsymbol{F}_{\text{ped}}^{(ij)} + \boldsymbol{F}_{\text{robot}}^{(i)} + \boldsymbol{F}_{\text{wall}}^{(i)} ,
\end{equation}
sets only the walking direction: pedestrian $i$ steers toward $\boldsymbol{F}_i/\|\boldsymbol{F}_i\|$ at its preferred speed, with an acceleration limit of 2\,m/s$^2$ per axis. $\boldsymbol{F}_{\text{goal}}^{(i)}$ points to the goal, and $\boldsymbol{F}_{\text{ped}}^{(ij)}$, $\boldsymbol{F}_{\text{robot}}^{(i)}$, and $\boldsymbol{F}_{\text{wall}}^{(i)}$ are repulsions from pedestrian $j$, the robot, and the walls that decay exponentially with distance. A pedestrian thus steers around the robot but does not slow down for it.

We add behaviors that make the crowd less predictable than the ORCA pedestrians of the standard benchmark. The preferred speed differs between pedestrians within $[0.6, 1.2]$\,m/s. Pedestrians stop at random (probability $p=0.0005$ per physics step, about 6\% per second, for 1--3\,s), which creates static obstacles that a purely velocity-based avoidance strategy handles poorly. Their reaction to the robot is \emph{asymmetric}: they respond to it only within a $200^\circ$ frontal arc and within 2.0\,m, whereas they respond to each other from any direction within the same range. This is deliberate: a person who has not seen or noticed the robot does not yield, so the policy can neither assume reciprocity nor rely on being avoided from afar. Pedestrians start at least 2.0\,m from the robot.

\subsection{Domain Randomization and Training Configuration}
\label{sec:randomization-config}

Instead of a staged curriculum, we randomize every episode: the crowd size, uniformly between 0 and 10; the robot start, 3.0\,m from the center of the circle at a random angle and heading; and the goal, at 3--7\,m within $\pm 90^\circ$ of the direction to the arena center and moved onto the 4.5\,m circle if it falls outside. Unused observation slots are zero except for $\text{TTC}_i = 1$ and are excluded from the attention statistics. Table~\ref{tab:hyperparams} lists the training configuration. Training ran for more than 5{,}300 iterations; we evaluate the checkpoint saved at iteration 2{,}500, the one nearest to the iteration with the highest training success rate. It was trained on $8\times10^{8}$ transitions, roughly 150 days of robot experience.

\begin{table}[tbp]
\caption{Training Configuration}
\label{tab:hyperparams}
{\centering\scriptsize
\begin{tabular}{|l|c|l|}
\hline
\textbf{Parameter} & \textbf{Value} & \textbf{Note}\\
\hline
Parallel environments & 10,000 & One L40S GPU (48\,GB)\\
\hline
Physics / control rate & 120 / 60\,Hz & Decimation 2\\
\hline
Episode length & 20\,s & 1{,}200 steps max\\
\hline
Steps per environment & 32 & Rollout length\\
\hline
Transitions per iteration & 320,000 & $10\text{k} \times 32$\\
\hline
Learning rate & $5 \times 10^{-5}$ & Initial; KL-adaptive\\
\hline
Discount $\gamma$ & 0.995 & $\approx$3.3\,s horizon\\
\hline
GAE $\lambda$ & 0.95 & ---\\
\hline
Entropy coefficient & 0.001 & Section~\ref{sec:ppo}\\
\hline
Evaluated iteration & 2,500 & $\approx$9\,h wall clock\\
\hline
\end{tabular}\par}
\end{table}

\subsection{ROS~2 Deployment}

The trained policy is wrapped in the \texttt{mipa\_rl\_planner} ROS~2 package. Positions and velocities are transformed into the body frame with TF2, and unused observation slots are filled with zeros and $\text{TTC}=1.0$ so that absent pedestrians read as no threat. The actor is evaluated deterministically, and a \texttt{speed\_scale} parameter scales down the commanded velocities for initial hardware runs.

Inference takes less than 10\,ms, and the loop runs at 60\,Hz to match training. The same node drives Gazebo and the physical robot, with an SFM crowd manager in place of pedestrian detection, and publishes the attention weights with each command for inspection at run time.

\section{EXPERIMENTS AND RESULTS}
\label{sec:results}

\subsection{Protocol and Metrics}
\label{sec:protocol}

The checkpoint at iteration 2{,}500, trained as described in Section~\ref{sec:randomization-config}, was evaluated at fixed crowd sizes of 5, 7, and 10 pedestrians, with deterministic (mean) actions and the observation noise of training. For each crowd size, we simulate 10{,}000 parallel environments, as in training, for 20.8\,s, just beyond the time limit, and count only the first episode of each environment. Every counted episode thus runs to its own end, and short episodes are not over-represented. The 95\% confidence intervals of the reported rates are within $\pm 1.0$ points.

We report five metrics. \emph{Success} is the fraction of episodes that reach the goal within the 20\,s limit without contact. \emph{Pedestrian collision} is the fraction terminated by $d_{\min} < r_{\text{combined}}$, and \emph{wall collision} the fraction terminated at the wall-distance limit, which the compared benchmarks lack. \emph{Timeout} is the fraction that reaches the time limit without any other termination, the standard proxy for freezing. \emph{Navigation time} is the mean duration of successful episodes.

\subsection{Randomized Versus Fixed Crowd Size}
\label{sec:randomization}

We trained a second policy with a fixed crowd of seven pedestrians; the code, reward, and training configuration are otherwise identical. Evaluated at the same iteration with the protocol of Section~\ref{sec:protocol}, it reaches 89.3\%, 86.2\%, and 79.1\% success with 5, 7, and 10 pedestrians, against 93.4\%, 89.2\%, and 78.8\% for the policy trained with randomized crowd sizes (Table~\ref{tab:results}): the randomized-crowd policy succeeds more often with five and seven pedestrians, and the two do not differ measurably with ten. The difference does not come from pedestrian avoidance. The fixed-crowd policy collides with pedestrians less often (3.5\%, 7.2\%, and 15.4\% against 4.5\%, 8.6\%, and 17.4\%), but it hits the walls more often in the smaller crowds (6.0\%, 5.1\%, and 3.6\% against 1.9\%, 2.2\%, and 3.6\%), times out in 1.3--2.0\% of the episodes instead of 0.3--0.4\%, and needs about 1\,s longer. The fixed-crowd variant was trained once, so this comparison is indicative.

\subsection{Performance Across Crowd Densities}

Table~\ref{tab:results} lists our results together with published values for the circle-crossing benchmark. Each method is quoted from its own paper, except CADRL and LSTM-RL, for which we quote the re-implementations in~\cite{chen2019crowd}. Methods evaluated under other protocols (SESN, Fl\"ogel et al., and DRL-VO) are omitted because their results are not comparable.

\begin{table}[tbp]
\caption{Navigation Performance}
\label{tab:results}
{\centering\scriptsize
\begin{tabular}{|l|c|c|c|c|c|c|}
\hline
\textbf{Method} & \textbf{Ped.} & \textbf{Succ.} & \textbf{P.\ col.} & \textbf{W.\ col.} & \textbf{T.out} & \textbf{Time}\\
\hline
CADRL~\cite{chen2017decentralized}$^\ast$ & 5 & 78.0 & 22.0 & --- & n/r & 10.80\\
\hline
LSTM-RL~\cite{everett2021collision}$^\ast$ & 5 & 95.0 & 3.0 & --- & n/r & 11.82\\
\hline
SARL~\cite{chen2019crowd} & 5 & 100.0 & 0.0 & --- & n/r & 10.55\\
\hline
RGL~\cite{chen2020relational} & 5 & 96.0 & 2.0 & --- & n/r & $^\ddagger$\\
\hline
SafeCrowdNav~\cite{xu2023safecrowdnav} & 5--10$^\S$ & 98.6 & 1.4 & --- & n/r & 9.98\\
\hline
MODSRL~\cite{cheng2024multi} & 5 & 99.8 & 0.2 & --- & 0.0 & 11.14\\
\hline
MODSRL~\cite{cheng2024multi} & 10 & 95.0 & 3.4 & --- & 1.6 & 14.13\\
\hline
MODSRL~\cite{cheng2024multi} & 15 & 87.4 & 7.8 & --- & 4.8 & 15.15\\
\hline
\textbf{\sys{}}$^\dagger$ & 5 & 93.4 & 4.5 & 1.9 & 0.3 & 8.91\\
\hline
\textbf{\sys{}}$^\dagger$ & 7 & 89.2 & 8.6 & 2.2 & 0.3 & 9.27\\
\hline
\textbf{\sys{}}$^\dagger$ & 10 & 78.8 & 17.4 & 3.6 & 0.4 & 9.72\\
\hline
\end{tabular}\par}
\vspace{1.2mm}
{\scriptsize All rates in \%, times in seconds, as reported in the cited papers ($^\ast$holonomic value-based re-implementations in~\cite{chen2019crowd}); n/r: not reported; ---: no walls. Baselines use the \emph{invisible-robot} circle-crossing protocol, in which pedestrians do not react to the robot. $^\ddagger$RGL reports only the extra time beyond the 8\,s optimum ($+1.86$\,s). $^\S$Scenes with five or ten pedestrians; the paper does not state which its table reports. $^\dagger$Measured under different conditions (non-holonomic kinematics, SFM pedestrians with stochastic stopping and asymmetric reaction, walls, goals at 3--7\,m instead of a fixed traversal) and not directly comparable; see text. Our rates can sum to over 100\% because about 0.1\% of the episodes meet two termination conditions in one step; our navigation times have a standard deviation of 2.3--2.8\,s.\par}
\end{table}

\textbf{Liveness.} The policy timed out in 29, 29, and 43 of the 10{,}000 episodes with 5, 7, and 10 pedestrians (0.3--0.4\%), with randomized start and goal positions. MODSRL, the only compared method that reports timeouts for several crowd sizes, reaches 1.6\% and 4.8\% with 10 and 15 pedestrians in the fixed environment of Table~\ref{tab:results}, and 12.8\% and 38.2\% in the random environment of the same study, with a 25\,s limit~\cite{cheng2024multi}; ours is 20\,s for routes of about 6\,m. TTC does not explain the low timeout rate: policies retrained without it time out as rarely (below 1.5\% in all eight at these crowd sizes). Disabling the selection in the trained policy does not raise the rate either (Section~\ref{sec:attribution}).

\textbf{Safety.} This liveness has a cost. Success falls from 93.4\% to 78.8\% between five and ten pedestrians, mainly because pedestrian collisions rise from 4.5\% to 17.4\%. With ten pedestrians, about one episode in six ends in a pedestrian collision. Logging the control step before each contact shows what these collisions are. With ten pedestrians, the pedestrian was approaching faster than the robot in 78\% of them (68\% with five), the robot was moving at less than 0.2\,m/s in 38\%, and the pedestrian was standing in only 7\%; in 94\%, the robot was inside the field of view of the pedestrian. Most collisions are thus pedestrians walking into a robot that they have seen, not the robot driving into people, and the rate depends on how weakly our simulated pedestrians yield as well as on the policy. A rule that commands zero linear velocity while the smallest TTC is below a threshold confirms this (Table~\ref{tab:components}): at 0.5\,s it removes only 6--15\% of the collisions, and at 2.0\,s it raises the collision rate with ten pedestrians to 30.9\% and the timeout rate to 14.0\%, because pedestrians walk into the stopped robot. We still consider the collision rate, not the timeout rate, the more important finding for deployment; Section~\ref{sec:limitations} discusses its causes and how it could be bounded.

\textbf{Comparability.} Table~\ref{tab:results} should not be read as a ranking. The baselines are evaluated with a robot that is invisible to the pedestrians, who therefore do not yield to it (in MODSRL, they respond only when in danger~\cite{cheng2024multi}); our pedestrians react within a $200^\circ$ frontal arc and ignore the robot otherwise. Our robot is also limited in its turn rate and cannot sidestep; our pedestrians are force-based and stop at random, and the motion model used in training changes the learned policy markedly~\cite{vandermeer2025comparative}; and our arena has walls, a failure mode absent from the baselines that accounts for 1.9--3.6 points of our failures. The navigation times are not comparable either, and normalizing by path length reverses their order: with five pedestrians, our robot travels about 6\,m on average to goals 3--7\,m away, whereas the MODSRL robot travels 8.78\,m~\cite{cheng2024multi}, which gives average speeds of $\approx$0.68\,m/s and $\approx$0.79\,m/s. Both robots are modeled as unicycles, so this is not a holonomic advantage, but the MODSRL action set offers eight orientations over the full circle per step, whereas our robot turns at up to 0.8\,rad/s and never reverses.

\subsection{Component Tests}
\label{sec:attribution}

Table~\ref{tab:components} also reports two tests in which we change a component of the policy of Table~\ref{tab:results} at test time, without retraining. With the selection thresholds disabled, so that both attention layers weight every pedestrian, no rate changes by more than 0.6 points: with up to ten pedestrians, the trained policy does not depend on discarding neighbors. With the TTC feature of every pedestrian set to 1 (no threat), success falls to 41.7\% with ten pedestrians, and the robot drives into pedestrians ahead of it at about 0.7\,m/s, so the policy relies on the TTC input for avoidance.

These tests show what the trained policy uses, not what training requires. For TTC, we retrained the policy for 2{,}500 iterations with four further seeds, each with and without TTC (input fixed at 1, reward term removed), and evaluated each, as in Section~\ref{sec:protocol}, at the checkpoint nearest to its highest training success rate. With 5, 7, and 10 pedestrians, the mean success rate is 92.4\%, 88.1\%, and 78.6\% with TTC against 88.2\%, 83.3\%, and 72.0\% without, and the mean pedestrian collision rate 4.7\%, 9.0\%, and 18.4\% against 8.6\%, 13.2\%, and 24.2\%; in all six comparisons, every seed with TTC is better than every seed without ($p = 0.029$, exact permutation test). At equal iterations, the mean success gain is 2--4, 2--6, and 4--7 points from iteration 1{,}500 on; earlier, one policy with TTC had not yet learned the task. Input and reward term are removed together, so their effects are not separated; the other components were not retrained.

\begin{table}[tbp]
\caption{Tests on the Trained Policy}
\label{tab:components}
{\centering\scriptsize
\begin{tabular}{|l|c|c|c|}
\hline
\textbf{Variant} & \textbf{Succ.} & \textbf{P.\ col.} & \textbf{T.out}\\
\hline
Unchanged & 93.4/89.2/78.8 & 4.5/8.6/17.4 & 0.3/0.3/0.4\\
\hline
Selection off & 93.0/88.8/78.4 & 4.9/8.7/17.9 & 0.2/0.3/0.3\\
\hline
TTC input $=1$ & 70.3/58.4/41.7 & 28.8/40.4/56.5 & 0.0/0.0/0.0\\
\hline
Stop if TTC $<0.5$\,s & 93.9/89.7/79.8 & 3.8/7.7/16.4 & 0.3/0.4/0.5\\
\hline
Stop if TTC $<1.0$\,s & 93.6/88.4/76.0 & 4.1/8.7/19.8 & 0.4/0.6/0.9\\
\hline
Stop if TTC $<1.5$\,s & 92.0/85.1/68.0 & 5.2/11.2/25.2 & 0.7/1.5/4.1\\
\hline
Stop if TTC $<2.0$\,s & 90.2/76.8/52.9 & 5.7/14.4/30.9 & 2.4/6.7/14.0\\
\hline
\end{tabular}\par}
\vspace{1.2mm}
{\scriptsize Rates in \% with 5 / 7 / 10 pedestrians, protocol of Section~\ref{sec:protocol}; the policy of Table~\ref{tab:results} is changed at test time only.\par}
\end{table}

\begin{figure}[tbp]
\centering
\includegraphics[width=0.9\columnwidth]{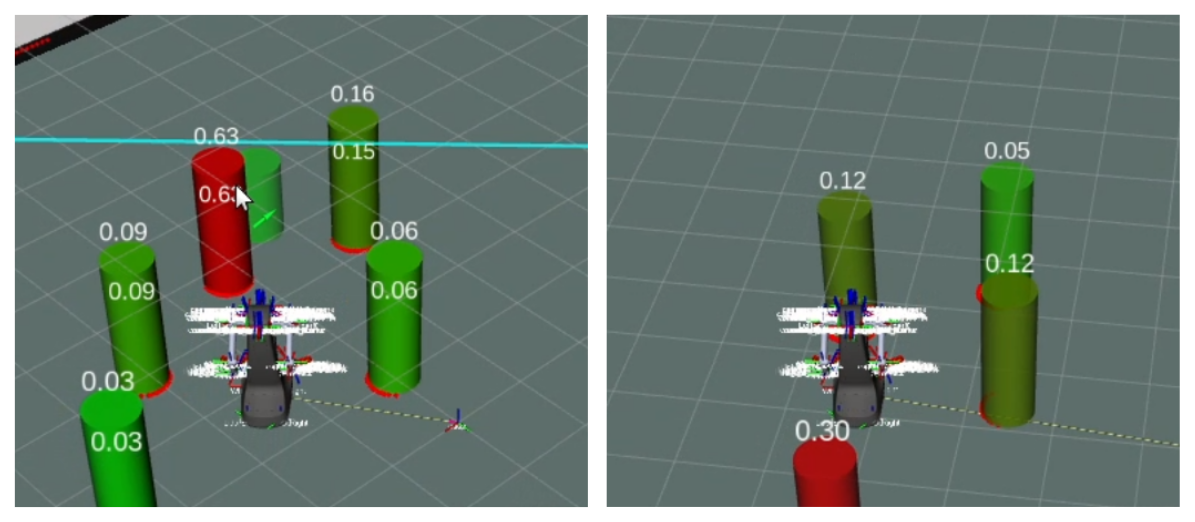}
\caption{Attention weights logged in Gazebo: first and last frames of a six-frame crossing sequence, with pedestrians colored by weight (red: high, green: low). The pedestrian that blocks the route to the goal receives $\approx$0.63, against 0.03--0.16 for the others; after the robot has crossed, the largest weight in the scene is $\approx$0.30.}
\label{fig:attention}
\end{figure}

\subsection{Sim-to-Sim and Hardware-in-the-Loop Validation}
\label{sec:hil}

The policy runs in Gazebo without retraining. The attention weights logged in Gazebo show the expected ranking (Fig.~\ref{fig:attention}): pedestrians ahead of the robot receive high weights, which fall once they are behind it.

On the physical MiPA robot, we ran a hardware-in-the-loop test. Localization (SLAM Toolbox), the ROS~2 node, and the drives are real, whereas the pedestrians are five virtual SFM agents injected into the map frame. No pedestrian perception is implemented on the robot: the people present during the test were not perceived by the robot. At the 60\,Hz control rate, with velocities scaled to 50\% for safety, the policy produced smooth trajectories that responded to the virtual pedestrians. This validates the deployment pipeline and the control interface end to end, but it is not a sim-to-real evaluation of navigation, which requires real, perceived pedestrians. No human subjects were involved in any experiment in this paper.

\section{DISCUSSION}

\subsection{Limitations}
\label{sec:limitations}

\textbf{Safety versus liveness.} The low timeout rate and the 17.4\% collision rate with ten pedestrians are most plausibly two consequences of the same incentives. Rewarding forward motion and penalizing a timeout as heavily as a collision make waiting unattractive, and so does our pedestrian model, in which a stopped robot is still walked into (Section~\ref{sec:results}). Among real people, a temporary stop is the safer failure, so the low timeout rate should be read as a property of this reward design and simulator, not as a deployment target. Bounding the collision rate while preserving progress requires pedestrians that yield realistically and an explicit constraint, such as constrained policy optimization or a control barrier function, rather than a reward weight; a stop rule alone is not enough.

\textbf{Attribution and protocol.} Retrained evidence exists only for TTC, with four seeds per variant; the other component tests use one trained policy, the fixed-crowd comparison rests on one run of that variant, and Table~\ref{tab:results} does not rank our system. Six of the eight retrained policies, three per variant, stand still when no pedestrian is present, although all eight navigate normally with one or more; the policy of Table~\ref{tab:results} does not show this defect. A controlled comparison under a single pedestrian model is still needed.

\textbf{Realism.} All experiments used virtual pedestrians, so the results say nothing about social acceptability. The circle-crossing scenario omits groups, sudden reversals, and narrow corridors; the observation model assumes reliable detection and tracking, which occlusion and sensor noise degrade in practice; and the observation holds at most 15 pedestrians.

\subsection{Toward Mobile Manipulation}

The target platform carries a manipulator, for which two properties of this controller are relevant. Continuous velocity commands with a penalty on fast rotation are meant to limit abrupt base motion, which disturbs a mounted arm, and the forward-only constraint keeps the sensors pointed in the direction of travel. Neither effect was measured.

\section{CONCLUSION}

This paper presented \sys{}, a navigation system for a commercial service robot that combines selective crowd attention, an anticipatory time-to-collision feature, and continuous non-holonomic control, trained in 10{,}000 parallel Isaac Lab environments and deployed as a ROS~2 node with sub-10\,ms inference. In 30{,}000 evaluation episodes, the policy timed out in fewer than 0.5\% at every tested crowd density, at the price of a pedestrian collision rate that reaches 17.4\% with ten pedestrians. We consider this trade-off, rather than the low timeout rate, the main result. Tests on the trained policy show that it relies on the TTC input but, with up to ten pedestrians, not on the selection thresholds; policies retrained without TTC reach a success rate that is 4--7 points lower.

Future work follows from Section~\ref{sec:limitations}: a controlled comparison with a continuous-action MODSRL, retrained ablations of the other components, and an explicit bound on the collision rate before any test with real people. Beyond that, we plan to integrate LiDAR-based detection and tracking, handle static obstacles through Nav2 costmaps, and evaluate denser crowds.

\section*{ACKNOWLEDGMENT}

Claude (Anthropic) was used to revise the wording of the entire manuscript and to check it against the implementation and the training logs. The authors reviewed all content and take full responsibility for it.

\bibliographystyle{IEEEtran}
\bibliography{references}

\begin{thebibliography}{10}
\providecommand{\url}[1]{#1}
\csname url@samestyle\endcsname
\providecommand{\newblock}{\relax}
\providecommand{\bibinfo}[2]{#2}
\providecommand{\BIBentrySTDinterwordspacing}{\spaceskip=0pt\relax}
\providecommand{\BIBentryALTinterwordstretchfactor}{4}
\providecommand{\BIBentryALTinterwordspacing}{\spaceskip=\fontdimen2\font plus
\BIBentryALTinterwordstretchfactor\fontdimen3\font minus
  \fontdimen4\font\relax}
\providecommand{\BIBforeignlanguage}[2]{{%
\expandafter\ifx\csname l@#1\endcsname\relax
\typeout{** WARNING: IEEEtran.bst: No hyphenation pattern has been}%
\typeout{** loaded for the language `#1'. Using the pattern for}%
\typeout{** the default language instead.}%
\else
\language=\csname l@#1\endcsname
\fi
#2}}
\providecommand{\BIBdecl}{\relax}
\BIBdecl

\bibitem{rahmati2026reinforcement}
\BIBentryALTinterwordspacing
M.~Rahmati, M.~Malmir, and A.~Knoll, ``Reinforcement learning with
  time-to-collision features and attention-based pedestrian filtering for crowd
  navigation,'' in \emph{RSS Workshop on Human-Centric Mobile Manipulation},
  2026, non-archival. [Online]. Available:
  \url{https://openreview.net/forum?id=SNCNojIbpW}
\BIBentrySTDinterwordspacing

\bibitem{trautman2010unfreezing}
P.~Trautman and A.~Krause, ``Unfreezing the robot: Navigation in dense,
  interacting crowds,'' in \emph{Proc. IEEE/RSJ Int. Conf. Intell. Robots Syst.
  (IROS)}, 2010, pp. 797--803.

\bibitem{le2024comprehensive}
H.~Le, S.~Saeedvand, and C.-C. Hsu, ``A comprehensive review of mobile robot
  navigation using deep reinforcement learning algorithms in crowded
  environments,'' \emph{J. Intell. Robot. Syst.}, vol. 110, no.~4, p. 158,
  2024.

\bibitem{chen2019crowd}
C.~Chen, Y.~Liu, S.~Kreiss, and A.~Alahi, ``Crowd-robot interaction:
  Crowd-aware robot navigation with attention-based deep reinforcement
  learning,'' in \emph{Proc. IEEE Int. Conf. Robot. Autom. (ICRA)}, 2019, pp.
  6015--6022.

\bibitem{chen2020relational}
C.~Chen, S.~Hu, P.~Nikdel, G.~Mori, and M.~Savva, ``Relational graph learning
  for crowd navigation,'' in \emph{Proc. IEEE/RSJ Int. Conf. Intell. Robots
  Syst. (IROS)}, 2020, pp. 10\,007--10\,013.

\bibitem{cheng2024multi}
C.-L. Cheng, C.-C. Hsu, S.~Saeedvand, and J.-H. Jo, ``Multi-objective
  crowd-aware robot navigation system using deep reinforcement learning,''
  \emph{Appl. Soft Comput.}, vol. 151, p. 111154, 2024.

\bibitem{everett2021collision}
M.~Everett, Y.~F. Chen, and J.~P. How, ``Collision avoidance in pedestrian-rich
  environments with deep reinforcement learning,'' \emph{IEEE Access}, vol.~9,
  pp. 10\,357--10\,377, 2021.

\bibitem{xie2023drlvo}
Z.~Xie and P.~Dames, ``{DRL-VO}: Learning to navigate through crowded dynamic
  scenes using velocity obstacles,'' \emph{IEEE Trans. Robot.}, vol.~39, no.~4,
  pp. 2700--2719, 2023.

\bibitem{flogel2024socially}
D.~Fl{\"o}gel, L.~Fischer, T.~Rudolf, T.~Sch{\"u}rmann, and S.~Hohmann,
  ``Socially integrated navigation: A social acting robot with deep
  reinforcement learning,'' in \emph{Proc. IEEE/RSJ Int. Conf. Intell. Robots
  Syst. (IROS)}, 2024, pp. 4785--4792.

\bibitem{neura2024mipa}
\BIBentryALTinterwordspacing
{NEURA Robotics}. {MiPA: My Intelligent Personal Assistant}. [Online].
  Available: \url{https://neura-robotics.com/products/mipa/}
\BIBentrySTDinterwordspacing

\bibitem{schulman2017proximal}
J.~Schulman, F.~Wolski, P.~Dhariwal, A.~Radford, and O.~Klimov, ``Proximal
  policy optimization algorithms,'' \emph{arXiv preprint arXiv:1707.06347},
  2017.

\bibitem{mittal2023orbit}
M.~Mittal, C.~Yu, Q.~Yu, J.~Liu, N.~Rudin, D.~Hoeller, J.~L. Yuan, R.~Singh,
  Y.~Guo, H.~Mazhar, A.~Mandlekar, B.~Babich, G.~State, M.~Hutter, and A.~Garg,
  ``Orbit: A unified simulation framework for interactive robot learning
  environments,'' \emph{IEEE Robot. Autom. Lett.}, vol.~8, no.~6, pp.
  3740--3747, 2023.

\bibitem{helbing1995social}
D.~Helbing and P.~Moln{\'a}r, ``Social force model for pedestrian dynamics,''
  \emph{Phys. Rev. E}, vol.~51, no.~5, pp. 4282--4286, 1995.

\bibitem{van2008reciprocal}
J.~van~den Berg, M.~Lin, and D.~Manocha, ``Reciprocal velocity obstacles for
  real-time multi-agent navigation,'' in \emph{Proc. IEEE Int. Conf. Robot.
  Autom. (ICRA)}, 2008, pp. 1928--1935.

\bibitem{vandenberg2011orca}
J.~van~den Berg, S.~J. Guy, M.~Lin, and D.~Manocha, ``Reciprocal $n$-body
  collision avoidance,'' in \emph{Robotics Research (ISRR)}, ser. Springer
  Tracts Adv. Robot.\hskip 1em plus 0.5em minus 0.4em\relax Springer, 2011, pp.
  3--19.

\bibitem{liu2023graph}
S.~Liu, P.~Chang, Z.~Huang, N.~Chakraborty, K.~Hong, W.~Liang, D.~L. McPherson,
  J.~Geng, and K.~Driggs-Campbell, ``Intention aware robot crowd navigation
  with attention-based interaction graph,'' in \emph{Proc. IEEE Int. Conf.
  Robot. Autom. (ICRA)}, 2023, pp. 12\,015--12\,021.

\bibitem{fox1997dynamic}
D.~Fox, W.~Burgard, and S.~Thrun, ``The dynamic window approach to collision
  avoidance,'' \emph{IEEE Robot. Autom. Mag.}, vol.~4, no.~1, pp. 23--33, 1997.

\bibitem{chen2017decentralized}
Y.~F. Chen, M.~Liu, M.~Everett, and J.~P. How, ``Decentralized
  non-communicating multiagent collision avoidance with deep reinforcement
  learning,'' in \emph{Proc. IEEE Int. Conf. Robot. Autom. (ICRA)}, 2017, pp.
  285--292.

\bibitem{xu2023safecrowdnav}
J.~Xu, W.~Zhang, J.~Cai, and H.~Liu, ``{SafeCrowdNav}: Safety evaluation of
  robot crowd navigation in complex scenes,'' \emph{Front. Neurorobot.},
  vol.~17, p. 1276519, 2023.

\bibitem{liu2023grrl}
Z.~Liu, Y.~Zhai, J.~Li, G.~Wang, Y.~Miao, and H.~Wang, ``Graph relational
  reinforcement learning for mobile robot navigation in large-scale crowded
  environments,'' \emph{IEEE Trans. Intell. Transp. Syst.}, vol.~24, no.~8, pp.
  8776--8787, 2023.

\bibitem{chen2017socially}
Y.~F. Chen, M.~Everett, M.~Liu, and J.~P. How, ``Socially aware motion planning
  with deep reinforcement learning,'' in \emph{Proc. IEEE/RSJ Int. Conf.
  Intell. Robots Syst. (IROS)}, 2017, pp. 1343--1350.

\bibitem{tian2025preference}
S.~Tian, C.~Wei, S.~Jian, and Z.~Ji, ``Preference-based deep reinforcement
  learning with automatic curriculum learning for map-free {UGV} navigation in
  factory-like environments,'' \emph{Eng. Sci. Technol. Int. J.}, vol.~70, p.
  102147, 2025.

\bibitem{zhu2026emobipednav}
W.~Zhu, A.~Raju, A.~Shamsah, A.~Wu, S.~Hutchinson, and Y.~Zhao,
  ``{EmoBipedNav}: Emotion-aware social navigation for bipedal robots with deep
  reinforcement learning,'' \emph{IEEE/ASME Trans. Mechatronics}, vol.~31,
  no.~5, pp. 5160--5172, 2026.

\bibitem{karamouzas2014universal}
I.~Karamouzas, B.~Skinner, and S.~J. Guy, ``Universal power law governing
  pedestrian interactions,'' \emph{Phys. Rev. Lett.}, vol. 113, no.~23, p.
  238701, 2014.

\bibitem{vandermeer2025comparative}
T.~Van Der~Meer, A.~Garulli, A.~Giannitrapani, and R.~Quartullo, ``A
  comparative study of human motion models in reinforcement learning algorithms
  for social robot navigation,'' \emph{ACM Trans. Hum.-Robot Interact.},
  vol.~15, no.~1, pp. 1--33, 2025.

\end{thebibliography}

\end{document}